\PassOptionsToPackage{table,xcdraw}{xcolor}
\documentclass[letterpaper]{article} 
\usepackage{ieee2026}  
\usepackage{times}  
\usepackage{helvet}  
\usepackage{courier}  
\usepackage[hyphens]{url}  
\usepackage{graphicx} 
\usepackage{natbib}  
\usepackage{caption} 
\usepackage{algorithm}
\usepackage{algorithmic}
\usepackage{soul}
\definecolor{mygray}{HTML}{EFEFEF}
\usepackage{multirow}
\usepackage{mathptmx}
\usepackage{amssymb}
\usepackage{newfloat}
\usepackage{listings}
\usepackage{booktabs}
\usepackage{xspace}
\definecolor{chenblue}{rgb}{0.21,0.49,0.74}
\usepackage[pagebackref,breaklinks,colorlinks,allcolors=chenblue]{hyperref}

\DeclareCaptionStyle{ruled}{labelfont=normalfont,labelsep=colon,strut=off} 
\floatstyle{ruled}
\newfloat{listing}{tb}{lst}{}
\floatname{listing}{Listing}

\title{Towards Unified Video Quality Assessment}

\author{
Chen Feng, Tianhao Peng, Fan Zhang, David Bull
}
\affiliations{
    Visual Information Lab, University of Bristol, BS1 5DD, United Kingdom\\
    \texttt{\{chen.feng, tianhao.peng, fan.zhang, dave.bull\}@bristol.ac.uk}
}

\begin{document}

\maketitle

\begin{abstract}
 Recent works in video quality assessment (VQA) typically employ monolithic models that typically predict a single quality score for each test video. These approaches cannot provide diagnostic, interpretable feedback, offering little insight into why the video quality is degraded. Most of them are also specialized, format-specific metrics rather than truly ``generic" solutions, as they are designed to learn a compromised representation from disparate perceptual domains. To address these limitations, this paper proposes \textbf{Unified-VQA}, a framework that provides a single, unified quality model applicable to various distortion types within multiple video formats by recasting generic VQA as a Diagnostic Mixture-of-Experts (MoE) problem. Unified-VQA employs  multiple ``perceptual experts'' dedicated to distinct perceptual domains. A novel multi-proxy expert training strategy is designed to optimize each expert using a ranking-inspired loss, guided by the most suitable proxy metric for its domain. We also integrated a diagnostic multi-task head into this framework to generate a global quality score and an interpretable multi-dimensional artifact vector, which is optimized using a weakly-supervised learning strategy, leveraging the known properties of the large-scale training database generated for this work. With static model parameters (without retraining or fine-tuning), Unified-VQA demonstrates consistent and superior performance compared to over 18 benchmark methods for both generic VQA and diagnostic artifact detection tasks across 17 databases containing diverse streaming artifacts in HD, UHD, HDR and HFR formats. This work represents an important step towards practical, actionable, and interpretable video quality assessment. 
\end{abstract}

\section{Introduction}
\label{sec:intro}

In the past decade, streamed video has become the predominant source of global internet traffic, delivering a diverse and escalating spectrum of user experiences, from High Definition (HD) video conferencing to immersive Ultra High Definition (UHD), High Frame Rate (HFR), and High Dynamic Range (HDR) entertainment \cite{r:cisco2020}. To monitor and maintain a high Quality of Experience (QoE), objective video quality assessment (VQA) has evolved from a research tool into a critical commercial necessity. It underpins the entire video delivery pipeline, from guiding perceptual encoding to enabling real-time online quality monitoring \cite{bull2021intelligent}.

\begin{figure}[!t]
    \centering
    \includegraphics[width=1\linewidth]{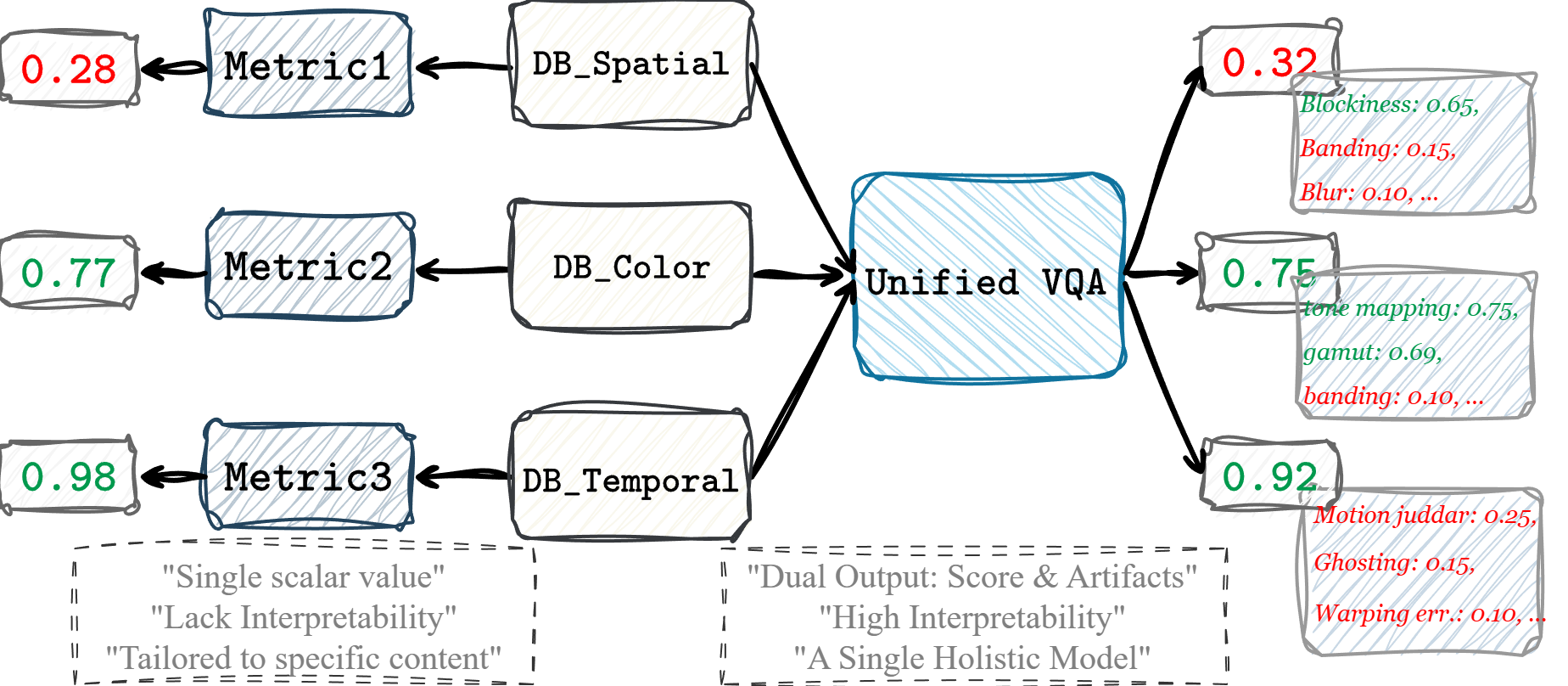}\vspace{15pt}

    \includegraphics[width=1\linewidth]{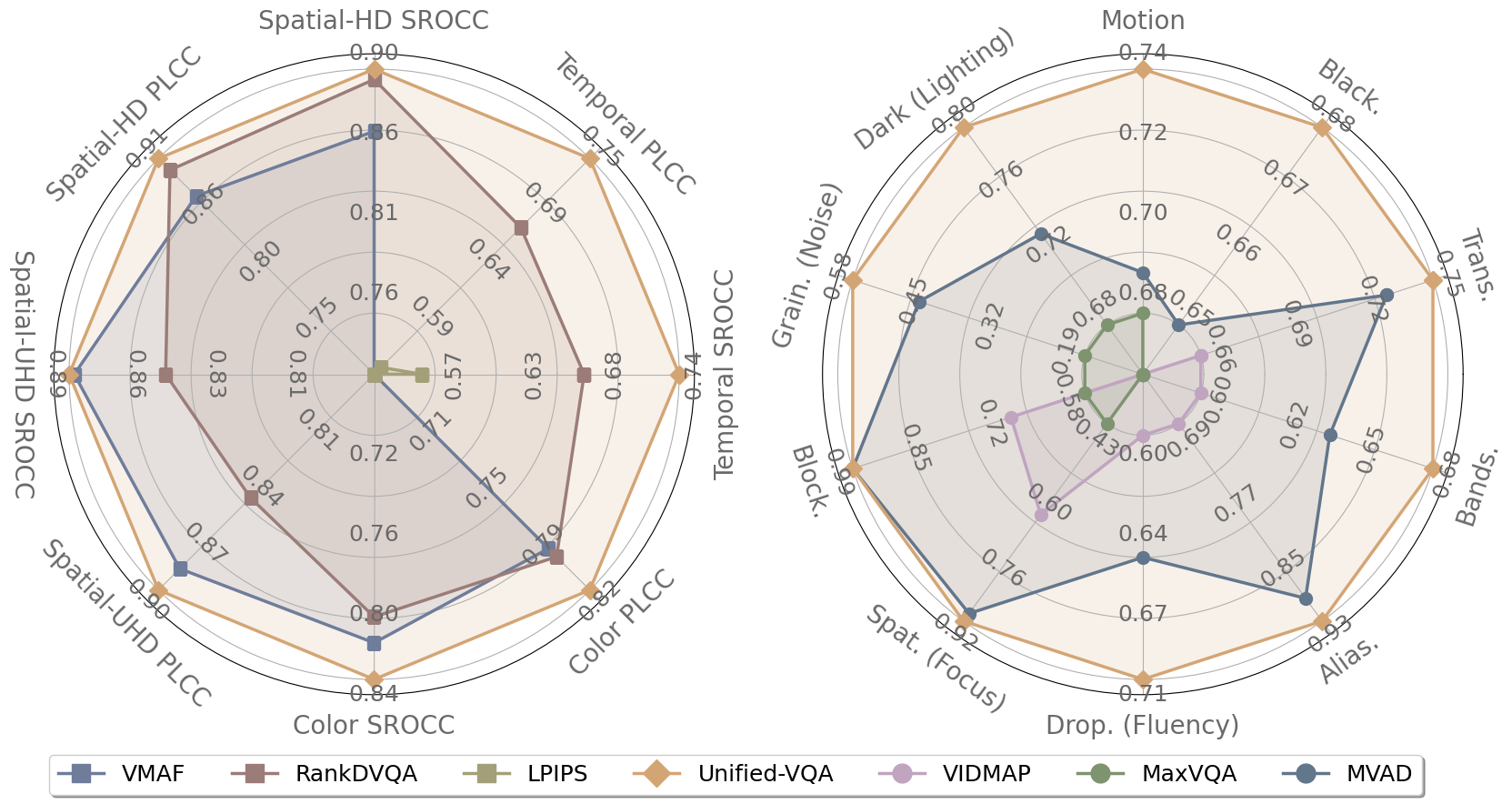}
    \caption{(Top) The difference between the proposed Unified-VQA framework and existing VQA models in terms of interpretability and practicality. (Bottom) Radar plots show the superior performance of Unified-VQA when compared to three well-performing quality models across four groups of databases containing compression artifacts (in HD and UHD formats), motion distortions and color artifacts for the VQA task (in SROCC and PLCC), when compared to three multi-artifact detectors (in F1-Score). Here large area indicates better overall performance.}
    \label{fig:highlight}
\end{figure}

However, the current literature in VQA is dominated by monolithic models that typically provide a single quality score for a given test video sequence \cite{saha2023perceptual, zheng2024video}. This paradigm has a critical limitation: it does not offer diagnostic and interpretable feedback - the overall quality score only tells service providers \textit{how good} a video is in general, with no insight  into \textit{why} it is degraded. A more fundamental issue is that these methods do not represent a \textit{generic} solution capable of handling diverse streaming artifacts across different video formats. Modern streaming is not confined to a single perceptual domain but rather constitutes a complex system with various spatial resolutions, frame rates, and dynamic ranges (bit depths), each of which introduces different types of artifacts. Forcing a single model to learn features for multiple distinct artifacts results in a compromised representation \cite{yu2020gradient} with suboptimal performance. Hence, the state-of-the-art solutions remain as specialized, format-specific metrics \cite{w:VMAF, c:Mantiuk, danier2022flolpips} rather than generic solutions. 

Recently, it has been demonstrated that VQA models can be optimized in a weakly-supervised manner using ranking-inspired losses and proxy labels \cite{hou2022perceptual, feng2024rankdvqa, rankiqa, conviqt}. However, these approaches still focus on producing a single score for each test video, without identifying or quantifying the existence of visual artifacts. While large vision-language models (VLLMs) for VQA do offer descriptive feedback, their significant computational cost often requires normalizing and downsampling diverse video formats \cite{wu2023qalign, li2025q}. This process makes it difficult to assess fine-grained artifacts specific to different streaming formats. Furthermore, the high inference cost of VLLMs makes them less practical for large-scale, real-time monitoring. It is also noted that multi-artifact detectors \cite{vidmap, tu2021ugc, feng2025mvad} do exist, which identify multiple co-existing artifacts in streaming videos; however, they cannot accurately quantify the impact of these artifacts on the overall perceptual quality score.

Based on these preliminary works, we propose \textbf{Unified-VQA} in this paper, converting generic VQA into a Diagnostic Mixture-of-Experts (MoE) problem, which aims to create a single, unified VQA model that explicitly disentangles disparate domains. Unified-VQA employs multiple ``perceptual experts'' dedicated to these distinct domains. To train these specialists, we introduce a novel multi-proxy expert training strategy with specialized SOTA quality metrics in each domain as guides, based on a ranking inspired loss. We then connect this backbone to a perceptually-inspired spatio-temporal aggregator, which fuses the extracted expert-specific features into a video representation. Finally, a diagnostic multi-task head generates two outputs based on the disentangled features from the experts: a high-accuracy global quality score and an interpretable, multi-dimensional artifact vector. This diagnostic head is optimized via a weakly-supervised artifact learning strategy, which leverages the known generation properties of our large-scale, multi-format database. The main contributions of this paper are summarized below:

\begin{itemize}    
    \item It is the first \textbf{unified VQA model} employing a novel Diagnostic Mixture-of-Experts (MoE) framework to disentangle visual artifacts across different perceptual domains, producing both global quality scores and artifact labels within a single model. This is fundamentally different from existing monolithic VQA models (shown in \autoref{fig:highlight} (top)).    
    \item The new \textbf{joint-training strategy}, for the first time, combines a multi-proxy expert ranking loss for global quality with a weakly-supervised artifact learning strategy for diagnostic interpretability.
    \item The benchmark and ablation study results demonstrate the ``\textbf{representational compromise}'' problem in monolithic VQA models, which in turn explains the fragmented state of the current VQA literature.
\end{itemize}

 Through an extensive experiment (based on 17 databases), the Unified-VQA framework consistently demonstrates superior performance over 18 public benchmarks for both generic VQA and diagnostic artifact detection tasks, without re-training/fine tuning the model, as shown in \autoref{fig:highlight} (bottom). As far as we know, this is the \textbf{first VQA model} achieving this level of performance, paving the way towards practical, actionable, and interpretable VQA solutions for multi-format video streaming services.

\section{Related Work}
\label{sec:related_work}

\noindent\textbf{Video Quality Assessment.} Early VQA methods (PSNR, SSIM \cite{ssim}) measure pixel-wise fidelity, which often do not fully align with human perception. To address this issue, perceptual quality models \cite{j:vsnr, MOVIE, PPVM} have been proposed to leverage the properties of the human visual system based on classic signal processing algorithms. These hand-crafted models can be further combined through regression, resulting in machine learning based approaches, e.g., VMAF \cite{w:VMAF}. More recently, based on advances in deep learning, end-to-end neural VQA models have been proposed using backbones such as 2D-CNN \cite{Kim_2017_CVPR, Kim2018DeepVQ}, LSTM \cite{VSFA, MDTVSFA}, 3D-CNN \cite{C3DVQA}, and hybrid CNN-Transformers \cite{starvqa, wu2022discovqa, FastVQA, feng2024rankdvqa}. These methods generally provide improved performance by learning spatio-temporal features directly from ground truth data,  typically requiring large training databases for model optimization. To alleviate this problem, ranking-inspired training methodologies have been proposed \cite{feng2024rankdvqa,hou2022perceptual}, which allow for the creation of large synthetic training sets with proxy quality labels annotated by existing quality models. However, these models remain monolithic \cite{{w:VMAF, C3DVQA, tu2021ugc}} - they do not support multiple video formats (e.g., SDR/HDR and UHD/HD) and cannot provide a unified solution for generic video quality assessment.

Beyond monolithic models, recent works have explored alternative paths to generalization. Ensemble methods like DOVER \cite{wu2023dover}, COVER \cite{he2024cover} and QualiVision~\cite{bompilwar2025qualivision} fuse the outputs of multiple existing VQA metrics to create a more robust, consensus-based score. While high-performing, these methods inherit the computational load of all their constituent metrics. Recently, the advent of Large Multimodal Models (LMMs) has enabled descriptive quality assessment \cite{wu2023qalign, ge2025lmm, li2025q, xing2025q}. They generate natural language explanations for quality issues, offering enhanced interpretability. However, their significant computational cost and the common need to downsample inputs make them less practical for large-scale monitoring and can obscure fine-grained, format-specific artifacts.

\begin{figure*}[!t]
\centering
\includegraphics[width=\textwidth]{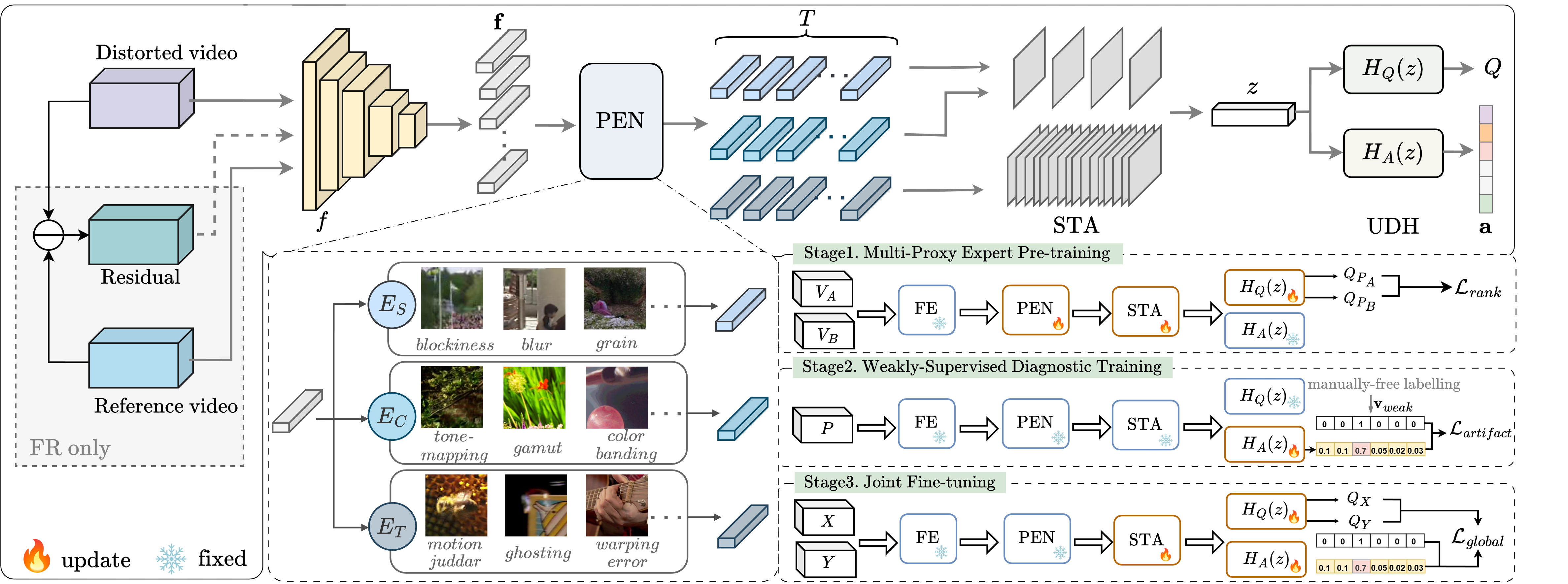}
    \caption{The proposed Unified-VQA framework, which consists of a pre-trained feature extractor $f$, three lightweight perceptual experts networks (PEN), a spatio-temporal aggregator (STA) and a unified diagnostic head. It has been trained based on an expert-guided multi-task learning strategy.}
\label{fig:architecture}
\end{figure*}

\vspace{5pt}  \noindent\textbf{Diagnostic Artifact Detection.} Existing VQA models lack interpretability, while some employ saliency maps \cite{zhang2020learning, li2022blindly} or vision-language models \cite{wu2023qalign, ge2025lmm} to explain quality scores, these typically provide either coarse regions or require heavy computation. Alternative works focus on building artifact detectors or classifiers that predict specific defects in streamed videos using supervised learning on labeled data \cite{wu2023towards,feng2024bvi}, yielding binary or continuous labels per artifact. However, these typically operate separately from VQA, for example, an artifact detector may flag ``blockiness'' but cannot determine whether the overall quality is good.

\section{Method}
\label{sec:method}

As illustrated in ~\autoref{fig:architecture}, the proposed Unified-VQA framework contains three primary components: (i) a Perceptual Expert Network (PEN) that disentangles feature extraction across three\footnote{\label{fn:K}The number of experts can be reconfigured for different applications. Here, we focus on three major aspects: spatial, temporal, and color distortions; hence, we employ three different expert networks.} independent aspects using domain-specific experts; (ii) a perceptually-inspired Spatio-Temporal Aggregator (STA) that fuses different expert features; and (iii) a Unified Diagnostic Head (UDH) that decomposes the output at the \textit{task} level, jointly predicting a global quality score and a diagnostic artifact vector. The entire model is optimized employing a novel expert-guided multi-task Learning strategy. 

The workflow of Unified-VQA begins with a shared 2D feature extractor ($f$), based on a pre-trained Vision Transformer (ViT-B/16) \cite{dosovitskiy2021vit}, that partitions the input video frames into non-overlapping patches and then linearly projects them into patch embeddings ($\mathbf{f} \in \mathbb{R}^{N \times D}$). These embeddings are then fed into the PEN module, which consists of three\textsuperscript{\ref{fn:K}} parallel, lightweight expert networks, each corresponding to a distinct perceptual domain. The features produced by these experts are collected and passed to the STA network, which is based on a SlowFast design that fuses the expert features across space and time into a single, comprehensive representation $z$. Finally, $z$ is fed into the UDH module, which outputs a single scalar $Q$ to predict the global quality score and an $N$-dimensional vector $\mathbf{a}$ to report the diagnostic artifacts.

This framework is designed to be flexible and can be instantiated as either a full-reference (FR) or a no-reference (NR) quality model. This divergence is handled at the input to the feature extractor. In the no-reference configuration, only the distorted video frames are processed, while in the full-reference configuration, the model takes both the distorted frames and the corresponding reference frames. Following \cite{Kim2018DeepVQ, feng2024rankdvqa}, these are concatenated together with their residual frames before being fed into the feature extractor. This allows the experts to model quality degradation relative to the reference content. In both configurations, the architectures of the STA and UDH modules, as well as the training methodology, remain the same.
 
\subsection{Perceptual Expert Network}
\label{ssec:peb}

The core hypothesis of this work is that training a single monolithic model on fundamentally disparate perceptual domains results in a \textit{representational compromise}, a form of `negative transfer' where competing task gradients can interfere with learning \cite{yu2020gradient}. For example, a single network based on the current methods trained to predict video quality on content with blockiness, high dynamic range tonal errors, and high frame rate interpolation artifacts will master neither\footnote{We will explicitly demonstrate this in our ablation studies (Section 5.4).}.

To address this problem, the PEN module employs a Mixture-of-Experts (MoE) design \cite{shazeer2017outrageously}. Unlike original MoE architectures \cite{fedus2022switch} that route tokens to generic, interchangeable experts to increase model capacity, the experts here are \textit{semantically distinct and specialized}. Each of the three experts is dedicated to a specific perceptual domain, corresponding to the major artifacts within a particular video format in modern streaming. Here, the feature embeddings $\mathbf{f}$ are fed into three\textsuperscript{\ref{fn:K}} parallel, lightweight adapter modules \cite{houlsby2019parameter} (the `experts'), as shown in ~\autoref{fig:architecture}, which are selected to represent three\textsuperscript{\ref{fn:K}} fundamental axes in the modern video delivery pipeline:

\begin{itemize}
    \item \textbf{$\texttt{Expert\_Spatial}$ ($E_S$)} learns features related to spatial-domain artifacts, such as blockiness, blur, and grain.
    
    \item \textbf{$\texttt{Expert\_Color}$ ($E_C$)} learns features related to color-domain artifacts, such as tone-mapping errors, color banding, and gamut correction distortions.
    
    \item \textbf{$\texttt{Expert\_Temporal}$ ($E_T$)} learns features related to temporal-domain artifacts, including motion judders, ghosting, and warping errors.
\end{itemize}

We implement these experts in a parameter-efficient manner using lightweight \textit{adapter modules} \cite{houlsby2019parameter} connected to the pre-trained shared feature extractor $f$. This allows each expert to learn a highly specialized, low-rank feature subspace while sharing the vast majority of computational parameters, making the model efficient to train and deploy. During training, batches from different domains are explicitly routed to their corresponding experts. This ensures that each expert learns features tailored to a specific perceptual domain. Unlike generic MoE, where experts are interchangeable, our experts have fixed roles, which removes the burden of forcing one model to learn across all domains simultaneously; this aims to solve the ``representational compromise'' problem with monolithic VQA \cite{yu2020gradient}.

\subsection{Spatio-Temporal Aggregator}

After obtaining expert-specific embeddings, we fuse them with a two-pathway SlowFast network, which was initially proposed for action recognition \cite{feichtenhofer2019slowfast} and later adapted for VQA \cite{sun2024enhancing}. This network processes inputs at a low frame-rate (slow path, rich spatial representation) and a high frame-rate (fast path, motion-focused) separately. Specifically, the outputs from the Spatial ($E_S$) and Color ($E_C$) experts are fed into the slow pathway to capture fine spatial/appearance cues, while the temporal features (from $E_T$) are passed to the Fast pathway to obtain temporal cues. Within the STA, the two pathways exchange information via lateral connections. Finally, following the original SlowFast design, the feature maps from both pathways undergo global average pooling. The resulting slow and fast feature vectors are then concatenated to form a single, comprehensive representation $z$, which is subsequently passed to the Unified Diagnostic Head.

\subsection{Unified Diagnostic Head}

The fused representation $z$ from STA is passed into two heads to generate the final output: (i) The global quality Head $H_Q(\cdot)$ is a regression head implemented as a Multi-Layer Perceptron (MLP) with two hidden layers, a GELU activation, and Dropout. It takes $z$ as input and maps it to a single output neuron with a linear activation. This allows the head to regress the feature $z$ to the global quality score $Q$. (ii) The diagnostic artifact head $H_A(\cdot)$ is a multi-label classification head, which is also implemented as an MLP that takes the same feature $z$ as input. Its final layer consists of $M$ output neurons, one for each artifact type being diagnosed. This layer is followed by a sigmoid activation function, producing the $N$-dimensional vector $\mathbf{a} \in [0, 1]^N$, where $N$ is the number of artifact types that the model is designed to recognize. Each element $a_i$ in this vector represents the predicted probability or severity of a specific artifact, enabling the model to provide interpretable diagnostic feedback. This dual-head design allows the model to perform both holistic quality scoring and fine-grained artifact diagnosis from the same shared representation.

\subsection{Expert-guided Multi-task Learning}
\label{ssec:training}

Multi-task learning (MTL) \cite{caruana1997multitask, yu2020gradient, zhang2021survey} jointly learns related tasks to improve robustness; however, for VQA, it has rarely been used beyond simple auxiliary outputs \cite{liebel2018auxiliary}. Recent works use mixture-of-experts (MoE) layers (gated expert networks) to scale NLP or vision models \cite{riquelme2021scaling, shazeer2017outrageously, fedus2022switch}, but usually for capacity, not interpretability. In this work, we propose a novel expert-guided multi-task learning strategy for training Unified-VQA. As shown in ~\autoref{fig:architecture}, this approach consists of three major stages: (i) a multi-proxy guidance strategy for optimizing the expert networks, the aggregator, and the global quality head; (ii) a weakly-supervised learning methodology for training the diagnostic head; and (iii) a joint loss to finetune the aggregator and the diagnostic head. This multi-stage training methodology creates a powerful synergy: the MoE backbone provides format-disentangled features ideal for the diagnostic head, while the auxiliary artifact loss reinforces the experts' specialization, further improving the global score $Q$. 

\vspace{5pt}\noindent\textbf{Stage 1: Multi-proxy expert pre-training.} 
The goal of this stage is to train the core quality assessment components: PEN ($E_S, E_C, E_T$), STA, and the global quality head ($H_Q$). This stage uses a large-scale dataset of patches with proxy labels to learn robust, domain-specific representations. We use pairs of distorted inputs ($V_A, V_B$) from a specific domain (e.g., spatial, color, or temporal). The model takes these videos and outputs two scalar quality scores, $Q_{A}$ and $Q_{B}$, from $H_Q$. We adapt the pairwise ranking loss from \cite{rankiqa, feng2024rankdvqa}. Each expert is guided by its best-suited proxy metric. For each pair, a binary label $V_{binary}$ is generated (where $V_{binary}=1$, if $V_A$ is better than $V_B$ according to its proxy; otherwise $V_{binary}=0$). The network is trained using a binary cross-entropy (BCE) loss, $\mathcal{L}_{rank}$:
\begin{equation}
\mathcal{L}_{rank} = - \mathrm{V}_{binary}\log(p) - (1 -\mathrm{V}_{binary} )\log(1-p).
\label{eq:s1}
\end{equation}
Here $p = \mathrm{sigmoid}(Q_{A} - Q_{B})$. This loss allows experts to learn their specializations without relying on expensive subjective tests.

\vspace{5pt}

\noindent\textbf{Stage 2: Weakly-supervised diagnostic training.}
This stage aims to train $H_A$ to detect specific artifacts. This is done while keeping the pre-trained backbone (PEN and STA) frozen. The model takes individual patches $P$ as input. The frozen backbone computes the representation $z$. The diagnostic head $H_A$ then outputs the artifact vector $\mathbf{a}$.

For the diagnostic head, acquiring dense, frame-level artifact annotations is prohibitively expensive \cite{feng2024bvi}. We overcome this by introducing a \textit{weakly supervised artifact learning} strategy, following \cite{feng2025mvad,shang2023study,hou2022perceptual}, which leverages the known properties of the training database generation to automatically produce a weak label vector $\mathbf{v}_{weak}$ for each patch (requiring zero manual annotation).

The artifact loss $\mathcal{L}_{artifact}$ is calculated as a Binary Cross-Entropy (BCE) loss between the predicted artifact vector $\mathbf{a}$ and the weak label $\mathbf{v}_{weak}$:
\begin{equation}
\mathcal{L}_{artifact} = \mathcal{L}_{BCE}(\mathbf{a}, \mathbf{v}_{weak}).
\label{eq:artifact}
\end{equation}
This loss enables the model to learn a feature representation $z$, which semantically understands the underlying causes of quality degradation.

\vspace{5pt}

\noindent\textbf{Stage 3: Joint fine-tuning on subjective scores.} The final stage fine-tunes STA and both prediction heads ($H_Q$ and $H_A$) jointly on a smaller dataset of \textit{full videos} with groundtruth subjective scores. The PEN expert parameters remain frozen to preserve their specialized knowledge. We use pairs of full video clips ($X, Y$) with their ground-truth subjective scores ($S_X, S_Y$). The model outputs two video-level quality scores ($Q_X, Q_Y$) from $H_Q$ and patch-level artifact vectors $\mathbf{a}$ from $H_A$. We first calculate the Global Quality Loss ($\mathcal{L}_{global}$) to align the model's predictions with human perception by operating on the \textit{difference} in scores \cite{feng2024rankdvqa}. Let $\delta_s = S_X - S_Y$ be the difference in subjective scores\footnote{The subjective scores are normalized to [0, 100].} and $\delta_q = Q_X - Q_Y$ be the difference in predicted scores. $\mathcal{L}_{global}$ is the L2 loss between these differences:
\begin{equation}
\mathcal{L}_{global} = \parallel \delta_q - \delta_s\parallel_2. 
\label{eq:global_s2}
\end{equation}
This loss is more robust than regressing on absolute scores and effectively teaches the model to learn the perceptual distance between videos.

The model is then trained with a combined multi-task loss $\mathcal{L}_{total}$:
\begin{equation}
    \mathcal{L}_{total} = \lambda_{g} \mathcal{L}_{global} + \lambda_{a} \mathcal{L}_{artifact},
\label{eq:total}
\end{equation}
in which $\lambda_{g}$ and $\lambda_{a}$ are hyperparameters balancing the two tasks. The artifact loss $\mathcal{L}_{artifact}$ is the same as in \autoref{eq:artifact}.

\section{Experiment}
\label{sec:experiment}

\vspace{5pt}\noindent\textbf{Training data generation.} Our model is optimized using a novel three-stage, expert-guided multi-task learning strategy. Each stage employs a distinct dataset tailored to its specific objective.

For Stage 1, we generated a large-scale, multi-domain dataset of video patch pairs. Rather than training all experts with a single, compromised proxy metric, we generated pseudo-labels for each perceptual domain using a specialized ``expert metric'' suitable for that task. This ensures that each expert learns a highly specialized and robust representation of its target domain. The sources and proxy metrics for each expert are summarized in \autoref{tab:training_data_experts}.There are approximately 270,800 patch pairs (with a size of 256$\times$256$\times$3$\times$12 \cite{feng2024rankdvqa}) used in this stage.

The generated training data is automatically annotated according to the following expert-proxy pairings:

\begin{itemize}
    \item Data for $E_S$: guided by VMAF~\cite{w:VMAF} or VMAF-4K~\cite{w:VMAF} (depending on the resolutions of the training batches).
    
    \item Data for $E_C$: guided by HDR-VDP-3~\cite{hdrvdp3}.
    
    \item Data for $E_T$: guided by VFIPS~\cite{hou2022perceptual}.
\end{itemize}

For Stage 2, we employ the large-scale artifact detection training database developed in ~\cite{feng2025mvad}. This dataset consists of approximately 50,880 video patches (e.g., $560 \times 560 \times 64$) derived from PGC sources (e.g., BVI-DVC~\cite{ma2021bvi}) and user-generated content (UGC) sources. Each patch is automatically annotated with a `weak' 10-dimensional binary label vector, indicating the presence or absence of ten common visual artifacts.

For Stage 3, we use a combined set of public VQA databases that provide reliable, sequence-level subjective scores. Following the methodology of ~\cite{feng2024rankdvqa}, we aggregate data from \textbf{VMAF+}~\cite{w:VMAF}, \textbf{IVP}~\cite{w:IVP}, and \textbf{VFIPS}~\cite{hou2022perceptual}. This combined dataset provides diverse examples of subjective quality across different content types and distortion domains.

\begin{table}[!t]
    \centering
    \caption{Training databases in expert pre-training). Data is partitioned by perceptual domain to train the corresponding expert.}\label{tab:training_data_experts}
    \resizebox{\linewidth}{!}{
    \begin{tabular}{rlc}
        \toprule
        \textbf{Expert} & \textbf{Content Sources} & \textbf{\# Patch Pairs} \\
        \midrule
        $E_S$ &  \begin{tabular}[t]{@{}l@{}}BVI-DVC~\cite{ma2021bvi},\\ CLIC~\cite{clic}, CableLabs~\cite{CableLabs} \\ (HD \& UHD SDR Content)\end{tabular} & $\sim$224,800 \\
        \midrule
        $E_C$ &  \begin{tabular}[t]{@{}l@{}}VTM HDR CTC~\cite{boyce2018jvet},\\ CableLabs~\cite{CableLabs} \\ (HDR Content)\end{tabular} & $\sim$20,000 \\
        \midrule
        $E_T$ & \begin{tabular}[t]{@{}l@{}}VFIPS~\cite{hou2022perceptual} \\ (HFR \& VFI Content)\end{tabular} & $\sim$26,000 \\
        \bottomrule
    \end{tabular}
    }
\end{table}

\vspace{5pt}\noindent\textbf{Evaluation tasks.}
To comprehensively evaluate the performance of the proposed Unified-VQA model, we target two important downstream tasks in a typical video streaming pipeline: video quality assessment and visual artifact detection. In our experiment, after the three-stage training process, the Unified-VQA model (both FR and NR variants) is frozen, and we do not perform any per-dataset fine-tuning, re-training, or cross-validation. This rigorous protocol measures the model's true out-of-distribution generalization \cite{wu2023dover, yuan2024ptm, duan2025finevq}. The same test configuration also applies to other learning-based quality models and artifact detectors.

\vspace{5pt}\noindent\textbf{Test datasets.}
For video quality assessment, we consider subjective databases containing three fundamental and disparate perceptual distortions: \textbf{spatial} artifacts (due to compression, resolution resampling, etc.), \textbf{color} distortions (due to HDR mapping errors, bit quantization, etc.), and \textbf{temporal} artifacts (due to motion judder, temporal down-sampling, and frame interpolation, etc.). For the first category, we also obtained databases with different spatial resolutions: HD and UHD. All these test databases are listed below:

\begin{itemize}
    \item \textbf{Spatial-HD} (8 databases): NFLX~\cite{w:VMAF}, NFLX-P~\cite{w:VMAF}, BVI-HD~\cite{zhang2018bvi}, CC-HD~\cite{bvi-cc}, CC-HDDO~\cite{c:Zhang24}, MCL-V~\cite{j:Lin4}, SHVC~\cite{r:JCTVCW0095} and VQEGHD3~\cite{r:vqegHD}

    \item \textbf{Spatial-UHD} (3 databases): AVT-VQDB-UHD~\cite{AVT-UHD}, Yonsei-UHD~\cite{yonseiUHD} and CC-UHD~\cite{bvi-cc}.
    
    \item \textbf{Color} (1 database): LIVE-HDR~\cite{shang2022subjective}.
    
    \item \textbf{Temporal} (3 databases): BVI-VFI~\cite{danier2022bvi}, LIVE-YT-HFR~\cite{livehfr}, and BVI-HFR~\cite{BVIhfr}
\end{itemize}

For artifact detection, following \cite{feng2025mvad}, we obtained two databases: Maxwell \cite{wu2023towards}\footnote{The Maxwell database contains content with eight artifact types, but we only test six of them that are relevant to video streaming.} and BVI-Artifact \cite{feng2024bvi}, focusing on ten common artifacts introduced in modern streaming pipelines, including motion blur, dark scenes, graininess, blockiness, spatial blur, dropped frames, aliasing, banding, transmission errors, and black frame.

\vspace{5pt}\noindent \textbf{Benchmark methods.} 
For video quality assessment, we compare Unified-VQA against a comprehensive set of benchmarks, including conventional full-reference (FR) metrics (PSNR, SSIM~\cite{ssim}, MS-SSIM~\cite{msssim}) and learning-based models (VMAF HD/4K~\cite{w:VMAF}, LPIPS~\cite{Lpips}, RankDVQA~\cite{feng2024rankdvqa}, VIDEVAL~\cite{tu2021ugc}, ChipQA~\cite{Chipqa}, and FastVQA~\cite{FastVQA}). We also compare against the bespoke HDR-VDP-3~\cite{hdrvdp3} metric on the LIVE-HDR dataset.
For artifact detection, following the practice in \cite{feng2025mvad}, we benchmark our diagnostic head against a suite of both single and multi-artifact detectors, including recent state-of-the-art models like CAMBI ~\cite{tandon2021cambi}, BBand~\cite{tu2020bband}, EFENet~\cite{zhao2021defocus}, MLDBD~\cite{zhao2023full},  Wolf \textit{et al.}~\cite{wolf2008no}, VIDMAP~\cite{vidmap}, MaxVQA~\cite{wu2023towards}, and MVAD~\cite{feng2025mvad}.

\vspace{5pt}\noindent\textbf{Evaluation Metrics}
For video quality assessment, following common practice, we report the Spearman Rank-Order Correlation Coefficient (SROCC). In contrast, for artifact detection, we evaluate the diagnostic task using F1 score.

\begin{table*}[!t]
\centering
\caption{Video quality assessment performance (and ablation study results) for the proposed Unified-VQA and other benchmark approaches. We also provided complexity figures including the number of model parameters and complexity (in GFLOPs).}
\label{tab:overall_plus_srocc}
\resizebox{0.95\linewidth}{!}{\begin{tabular}{r|c|c|c|c| ccc}
\toprule
\textbf{Average SROCC} ↑ & \textbf{Spatial-HD (8)} & \textbf{Spatial-UHD (3)} & \textbf{Color (1)} &  \textbf{Temporal (3)} & \textbf{Parameters (M)} & \textbf{GFLOPs} \\
\midrule 
\multicolumn{5}{c}{\textbf{Full Reference (FR) Methods}} \\
\midrule
PSNR              & 0.6520$\pm$0.0970    & 0.6798$\pm$0.1268   & 0.3760 & 0.5651$\pm$0.1899 &--- &---\\
SSIM  \cite{ssim} & 0.6216$\pm$0.1275    & 0.7236$\pm$0.1183   & 0.6976   & 0.4438$\pm$0.1609 &--- &---\\
MS-SSIM \cite{msssim} & 0.7601$\pm$0.0693    & 0.7946$\pm$0.0819   & 0.6757   & 0.4774$\pm$0.1529 &--- &---\\
VMAF HD/4K \cite{w:VMAF} & 0.8644$\pm$0.0549    & 0.8882$\pm$0.0415    & 0.8184   & 0.5170$\pm$0.3004 &--- &---\\
LPIPS \cite{Lpips} & 0.7107$\pm$0.0658    & 0.7792$\pm$0.0345   & 0.6826   & 0.5512$\pm$0.1275 &2.47 &20.86\\
HDR-VDP-3  \cite{hdrvdp3}   &    ---         &   ---               & 0.8080   &   ---  &--- &---\\
RankDVQA (FR) \cite{feng2024rankdvqa}     & 0.8972$\pm$0.0314     & 0.8550$\pm$0.0309 & 0.8053   & 0.6675$\pm$0.0743 & 25.6 & 39.5\\
\midrule
\textbf{Unified-VQA (FR)} & \textbf{0.9037$\pm$0.0323} & \textbf{0.8901$\pm$0.0287} &\textbf{0.8369}& \textbf{0.7356$\pm$0.0731} & 31.8  & 48.2 \\
\midrule
\textbf{V1} (Single expert Net)& 0.8829$\pm$0.0362 & 0.8628$\pm$0.0314 & 0.8162 & 0.7003$\pm$0.0746 & 31.0  & 47.1 \\
\textbf{V2} (CNN Aggregator) & 0.8988$\pm$0.0353 & 0.8785$\pm$0.0307 & 0.8194 & 0.6732$\pm$0.1385 & 27.8  & 40.9 \\
\textbf{V3} (Without UDH) & 0.9002$\pm$0.0318 & 0.8846$\pm$0.0299 & 0.8208 & 0.7124$\pm$0.0912 & 28.4  & 43.2 \\
\textbf{V4} ($E_S$ HD data)& 0.8976$\pm$0.0346 & 0.8673$\pm$0.0311 & 0.8095 & 0.6746$\pm$0.0789 & 31.8  & 48.2 \\
\textbf{V5} (full $E_S$ data)& 0.9018$\pm$0.0337 & 0.8854$\pm$0.0315 & 0.8115 & 0.6734$\pm$0.1327& 31.8  & 48.2 \\
\textbf{V6} ($E_S$ + $E_C$ data) & 0.9025$\pm$0.0329 & 0.8894$\pm$0.0297 & 0.8361 & 0.6982$\pm$0.0921 & 31.8  & 48.2 \\
\textbf{V7} ($E_S$ + $E_C$ + $E_T$ data)& 0.9037$\pm$0.0323 & 0.8901$\pm$0.0287 & 0.8369 & 0.7356$\pm$0.0731 & 31.8  & 48.2 \\
\midrule 
\multicolumn{5}{c}{\textbf{No Reference (NR) Methods}} \\
\midrule
VIDEVAL \cite{tu2021ugc}     & 0.6529$\pm$0.1105      &   0.5728$\pm$0.1029            & 0.4187   & 0.3007$\pm$0.1993 &--- &---\\
ChipQA  \cite{Chipqa}   & 0.6829$\pm$0.0692      & 0.6140$\pm$0.0596 & 0.7435   & 0.3722$\pm$0.0565 &--- &---\\
FastVQA  \cite{FastVQA}         &    0.7164$\pm$0.08362                & 0.5716$\pm$0.0999 & 0.6538   & 0.2952$\pm$0.1367 &27.7 &279\\
RankDVQA (NR) \cite{feng2024rankdvqa}    & 0.7791$\pm$0.0448     & 0.7616$\pm$0.0284 & 0.7108   & 0.5568$\pm$0.0950 & 24.9 & 24.6 \\
\midrule
\textbf{Unified-VQA (NR)} & \textbf{0.7987$\pm$0.0426} & \textbf{0.7853$\pm$0.0311}& \textbf{0.7524} & \textbf{0.6523$\pm$0.0724}  & 31.2  & 28.7\\
\bottomrule
\end{tabular}}
\end{table*}

\vspace{5pt}\noindent\textbf{Implementation Details.} We segment full-length videos into non-overlapping 12-frame clips. Each 12-frame clip is passed to the feature extractor, yielding a clip-level quality score $Q_{clip}$ and an artifact vector $\mathbf{a}_{clip}$. The final video-level quality score $Q_{video}$ is the temporal average of all $Q_{clip}$ scores. The corresponding $\mathbf{a}_{clip}$ vectors are concatenated to form a video-wide diagnostic map of artifact presence. This clip-based approach scales to longer clips, given hardware with sufficient memory. Other implementation configurations are provided as follows. Pytorch 1.10 was used to implement Unified-VQA, with the following training parameters: Adam optimization~\cite{kingma2014adam} with $\beta_1$=0.9 and $\beta_2$=0.999; 60 training epochs; a batch size of 4; the initial learning rate is 0.0001 with a weight decay of 0.1 after every 20 epochs. The backbone of 2D feature extractor ($f$) employs the Vision Transformer (ViT-B/16) \cite{dosovitskiy2021vit} with weights pre-trained on ImageNet-21k \cite{deng2009imagenet}. Both training and evaluation were executed on a computer with a 2.4GHz Intel CPU and an NVIDIA P100 GPU.

\section{Results and Discussion}
\label{sec:results}

\subsection{Video Quality Assessment Performance}
\label{sec:quant_plus}

\autoref{tab:overall_plus_srocc} summarizes the VQA results of the proposed Unified-VQA (both FR and NR versions) and their benchmark methods. Here, for each group of test datasets: Spatial-HD, Spatial-UHD, Color, and Temporal, we report the mean and standard deviation of SROCC values among the databases. It can be observed that for all four groups, Unified-VQA (FR) consistently offers better correlation performance with the subjective ground truth compared to all the benchmarks. Its no-reference version, Unified-VQA (NR), also outperforms its corresponding benchmarks for each group of datasets. This highlights the superior model generalization ability of the proposed method. The second best performing models are RankDVQA (for Spatial-HD), VMAF-4K (for Spatial-UHD), VMAF-4K (for Color) and RankDVQA (temporal). More detailed results for each test dataset are provided in the \textit{Supplementary}. 

\begin{figure}[!t]
    \centering
    \includegraphics[width=1\linewidth]{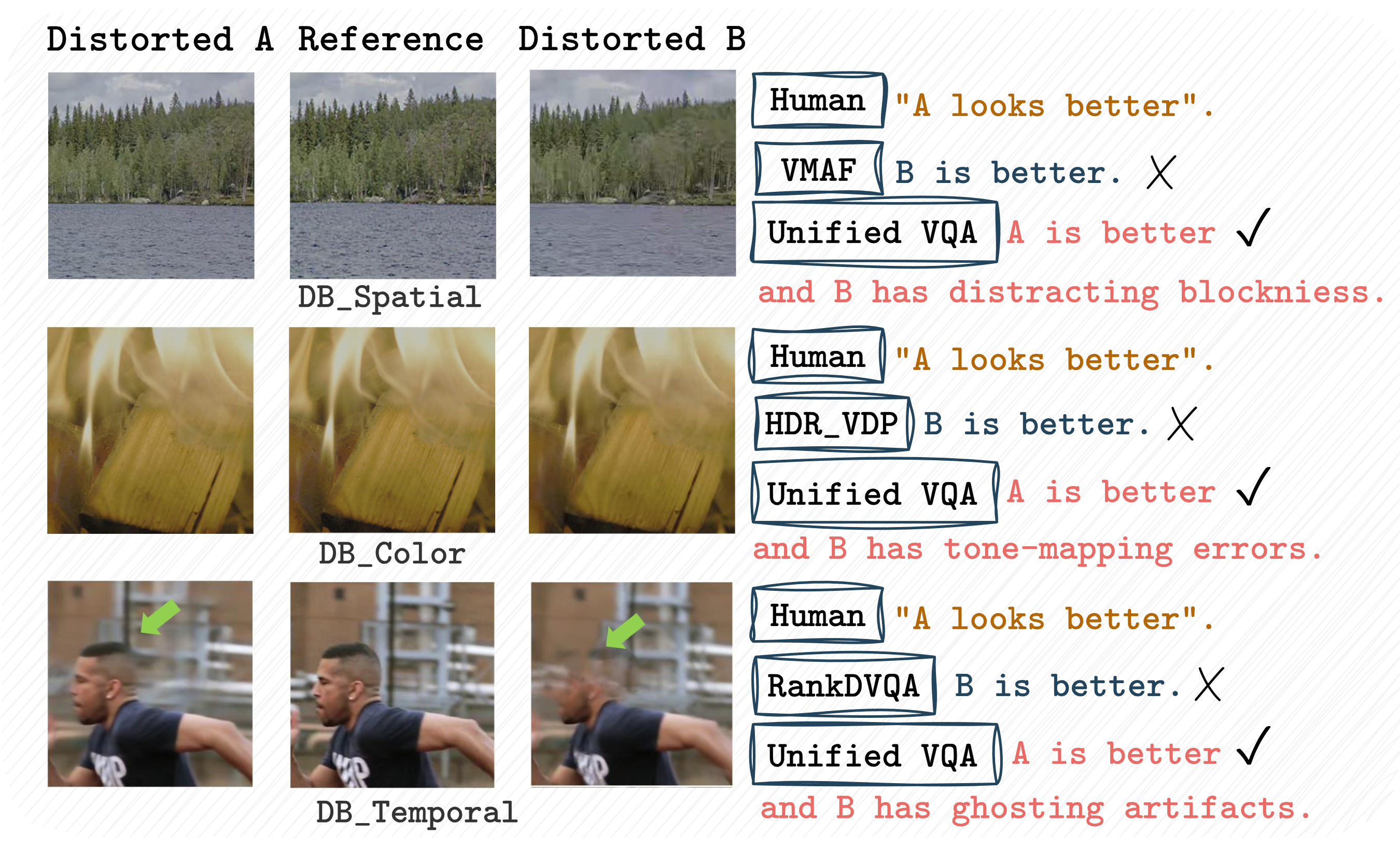}
    \caption{Qualitative comparison between Unified-VQA (FR) and existing quality models, which shows that Unified-VQA aligns more closely with the human decisions for each content group.}
    \label{fig:qualitative_examples_chap5}
\end{figure}

Qualitative results are also provided in \autoref{fig:qualitative_examples_chap5}, which show that Unified-VQA offers more accurate quality ranking decisions compared to benchmarking methods for each content group.

\subsection{Artifact Detection Performance}

\begin{table*}[t!]
\caption{Artifact detection F1 results on the Maxwell and BVI-Artefact databases. Here `\colorbox{mygray}{ - }' indicates that the tested method in this row is not designed to identify the corresponding artifact in this column or the corresponding dataset does not contain that artifact.}
    \label{tab:artifact_f1}
    \centering
    \resizebox{0.99\linewidth}{!}{%
    \begin{tabular}{c r c c c c c c c c c c}
    \toprule
       Database & Method  & Motion & Dark (Lighting) & Grain.(Noise) & Block. (Compression) & Spat.(Focus) & Drop. (Fluency) & Alias. & Bands. & Trans. &  Black.\\
    \midrule
    \multirow{6}{*}{\rotatebox{270}{Maxwell}} 
      & EFENet \cite{zhao2021defocus} 
        & \cellcolor[HTML]{EFEFEF}-         & \cellcolor[HTML]{EFEFEF}-         & \cellcolor[HTML]{EFEFEF}-         & \cellcolor[HTML]{EFEFEF}-         & 0.66         & \cellcolor[HTML]{EFEFEF}-  & \cellcolor[HTML]{EFEFEF}-  & \cellcolor[HTML]{EFEFEF}-  & \cellcolor[HTML]{EFEFEF}-  & \cellcolor[HTML]{EFEFEF}- \\
      & MLDBD \cite{zhao2023full} 
        & \cellcolor[HTML]{EFEFEF}-         & \cellcolor[HTML]{EFEFEF}-         & \cellcolor[HTML]{EFEFEF}-         & \cellcolor[HTML]{EFEFEF}-         & 0.63         & \cellcolor[HTML]{EFEFEF}-   & \cellcolor[HTML]{EFEFEF}-  & \cellcolor[HTML]{EFEFEF}-  & \cellcolor[HTML]{EFEFEF}-  & \cellcolor[HTML]{EFEFEF}- \\
      & Wolf \textit{et al.} \cite{wolf2008no} 
        & \cellcolor[HTML]{EFEFEF}-         & \cellcolor[HTML]{EFEFEF}-         & \cellcolor[HTML]{EFEFEF}-         & \cellcolor[HTML]{EFEFEF}-         & \cellcolor[HTML]{EFEFEF}-         & 0.59  & \cellcolor[HTML]{EFEFEF}-  & \cellcolor[HTML]{EFEFEF}-  & \cellcolor[HTML]{EFEFEF}-  & \cellcolor[HTML]{EFEFEF}- \\
      & VIDMAP \cite{vidmap} 
        & \cellcolor[HTML]{EFEFEF}-         & \cellcolor[HTML]{EFEFEF}-         & \cellcolor[HTML]{EFEFEF}-         & 0.71         & 0.64         & 0.62   & \cellcolor[HTML]{EFEFEF}-  & \cellcolor[HTML]{EFEFEF}-  & \cellcolor[HTML]{EFEFEF}-  & \cellcolor[HTML]{EFEFEF}- \\
      & MaxVQA \cite{tu2021ugc} 
        & 0.75         & 0.72         & 0.66         & 0.85         & 0.82         & 0.77   & \cellcolor[HTML]{EFEFEF}-  & \cellcolor[HTML]{EFEFEF}-  & \cellcolor[HTML]{EFEFEF}-  & \cellcolor[HTML]{EFEFEF}- \\
      & MVAD \cite{feng2025mvad} 
        & 0.78         & 0.80         & 0.72         & \textbf{0.99}        & \textbf{0.92}         & \textbf{0.79}  & \cellcolor[HTML]{EFEFEF}-  & \cellcolor[HTML]{EFEFEF}-  & \cellcolor[HTML]{EFEFEF}-  & \cellcolor[HTML]{EFEFEF}- \\ 
                \cmidrule{2-12}
      & \textbf{Unified-VQA} 
        & \textbf{0.80}         & \textbf{0.82}         & \textbf{0.75}        & \textbf{0.99}         & \textbf{0.92}         & \textbf{0.79}   & \cellcolor[HTML]{EFEFEF}-  & \cellcolor[HTML]{EFEFEF}-  & \cellcolor[HTML]{EFEFEF}-  & \cellcolor[HTML]{EFEFEF}- \\
    \midrule
    \multirow{9}{*}{\rotatebox{270}{BVI-Artefact}} 
    & CAMBI \cite{tandon2021cambi} & \cellcolor[HTML]{EFEFEF}-& \cellcolor[HTML]{EFEFEF}-& \cellcolor[HTML]{EFEFEF}-& \cellcolor[HTML]{EFEFEF}-& \cellcolor[HTML]{EFEFEF}-& \cellcolor[HTML]{EFEFEF}- & \cellcolor[HTML]{EFEFEF}- & 0.53 & \cellcolor[HTML]{EFEFEF}- & \cellcolor[HTML]{EFEFEF}-\\
    & BBAND \cite{tu2020bband} & \cellcolor[HTML]{EFEFEF}-& \cellcolor[HTML]{EFEFEF}-& \cellcolor[HTML]{EFEFEF}-& \cellcolor[HTML]{EFEFEF}-& \cellcolor[HTML]{EFEFEF}-& \cellcolor[HTML]{EFEFEF}- & \cellcolor[HTML]{EFEFEF}- & 0.44 & \cellcolor[HTML]{EFEFEF}- & \cellcolor[HTML]{EFEFEF}-\\
      & EFENet \cite{zhao2021defocus} 
        & \cellcolor[HTML]{EFEFEF}-         & \cellcolor[HTML]{EFEFEF}-         & \cellcolor[HTML]{EFEFEF}-         & \cellcolor[HTML]{EFEFEF}-         & 0.64         & \cellcolor[HTML]{EFEFEF}- & \cellcolor[HTML]{EFEFEF}- & \cellcolor[HTML]{EFEFEF}- & \cellcolor[HTML]{EFEFEF}- & \cellcolor[HTML]{EFEFEF}-\\
      & MLDBD \cite{zhao2023full} 
        & \cellcolor[HTML]{EFEFEF}-         & \cellcolor[HTML]{EFEFEF}-         & \cellcolor[HTML]{EFEFEF}-         & \cellcolor[HTML]{EFEFEF}-         & 0.65         & \cellcolor[HTML]{EFEFEF}- & \cellcolor[HTML]{EFEFEF}- & \cellcolor[HTML]{EFEFEF}- & \cellcolor[HTML]{EFEFEF}- & \cellcolor[HTML]{EFEFEF}-\\
      & Wolf \textit{et al.} \cite{wolf2008no} 
        & \cellcolor[HTML]{EFEFEF}-         & \cellcolor[HTML]{EFEFEF}-         & \cellcolor[HTML]{EFEFEF}-         & \cellcolor[HTML]{EFEFEF}-         & \cellcolor[HTML]{EFEFEF}-         & 0.18 & \cellcolor[HTML]{EFEFEF}- & \cellcolor[HTML]{EFEFEF}- & \cellcolor[HTML]{EFEFEF}- & \cellcolor[HTML]{EFEFEF}-\\
      & VIDMAP \cite{vidmap} 
        & \cellcolor[HTML]{EFEFEF}-         & \cellcolor[HTML]{EFEFEF}-         & \cellcolor[HTML]{EFEFEF}-         & 0.69         & 0.64         & 0.59  & 0.67 & 0.59 & 0.65 & \cellcolor[HTML]{EFEFEF}-\\
      & MaxVQA \cite{tu2021ugc}         & 0.68         & 0.67         & 0.16         & 0.55         & 0.40         & \cellcolor[HTML]{EFEFEF}- & \cellcolor[HTML]{EFEFEF}- & \cellcolor[HTML]{EFEFEF}- & \cellcolor[HTML]{EFEFEF}- & \cellcolor[HTML]{EFEFEF}- \\
      & MVAD \cite{feng2025mvad}
        & 0.69         & 0.73         & 0.46       & \textbf{0.99}       & 0.90         & 0.65  & 0.90 & 0.64 & 0.73 & 0.65\\
        \cmidrule{2-12}
      & \textbf{Unified-VQA} 
        & \textbf{0.74}         & \textbf{0.80}         & \textbf{0.58}         & \textbf{0.99}        & \textbf{0.92}         & \textbf{0.71}  & \textbf{0.93}  & \textbf{0.68}  & \textbf{0.75} & \textbf{0.68}\\
    \bottomrule
    \end{tabular}}
\end{table*}

Table~\ref{tab:artifact_f1} reports the artifact detection results (as F1-scores) of the proposed method and other benchmark approaches. It can be observed that Unified-VQA outperforms multi-artifact detectors, MaxVQA~\cite{wu2023towards} and MVAD~\cite{feng2025mvad}, and all single artifact models. These results demonstrate the interpretability of the proposed method, which is important for a practical quality monitoring tool. More detailed results based on other accuracy metrics are provided in the \textit{Supplementary}.

\subsection{Model Complexity}
\label{sec:complexity}

The complexity figures for Unified-VQA are provided in~\autoref{tab:overall_plus_srocc}, which shows that our FR model is associated with 31.8M parameters and 48.2 GFLOPs. These are slightly higher than those for the non-diagnostic quality model, RankDVQA. 

\subsection{Ablation Studies}
\label{sec:ablation}

To validate the contribution of each component in our framework. We conducted ablation studies and report the results in \autoref{tab:overall_plus_srocc}. Here, we solely focus on the full reference model in this ablation study and obtain results for the VQA task. More ablation study results can be found in the \textit{Supplementary}.

\vspace{5pt}\noindent\textbf{Effectiveness of the Perceptual Expert Network (PEN).} We implemented (V1) to replace our Perceptual Expert Networks with a single, monolithic backbone of equivalent parameter count. This change causes a significant drop in performance for VQA, as shown in \autoref{tab:overall_plus_srocc}. This validates our core hypothesis in this paper: the monolithic model suffers from ``representational compromise'', failing to learn across multiple perceptual domains. It also confirms that the PEN architecture is crucial for disentangling these domains and achieving expert performance in all of them.

\vspace{5pt}\noindent\textbf{Effectiveness of the Spatio-Temporal Aggregator (STA).}
We created V2 by replacing our SlowFast-based STA with the simpler convolutional temporal aggregator from the original RankDVQA~\cite{feng2024rankdvqa}. While VQA performance on the spatial-HD content is slightly affected, the performance on other database groups drops significantly, particularly for Temporal content. This validates our choice of the SlowFast architecture as a superior method for modeling complex spatio-temporal dynamics, which is essential for temporal artifacts.

\vspace{5pt}\noindent\textbf{Synergy of the Unified Diagnostic Head (UDH) \& Loss.}
We also created V3 to remove the Unified Diagnostic Head ($H_A$) and its associated artifact loss ($\mathcal{L}_{artifact}$), training the model for VQA only. This not only disables the model's diagnostic capability but also degrades the primary VQA performance, as shown in \autoref{tab:overall_plus_srocc}. This reveals that multi-task learning, guided by the $\mathcal{L}_{artifact}$ loss, acts as a strong regularizer. It enables the PEN backbone to learn more robust, disentangled, and semantically meaningful features, which, in turn, benefits the global quality prediction.

\vspace{5pt}\noindent\textbf{Importance of Multi-Format Training Data.}
V4-V7 are trained with reduced training content (the number of experts is reduced as well) as shown in \autoref{tab:overall_plus_srocc}. As expected, all these variants achieve lower correlation performance across all groups of databases compared to the original Unified-VQA. This confirms that multiple experts trained on diverse, training material is essential for building a truly generic and generalizable VQA model.

\section{Conclusion}
\label{sec:conclusion}
We have presented Unified-VQA, a unified and expert-guided framework for generic video quality assessment (VQA) across diverse formats, including HD, UHD, HDR, and HFR/VFI. Our method extends the RankDVQA foundation with three key innovations: (1) a modular expert design with domain-specialized heads for content and artifact modeling, (2) a biologically inspired STSFNet aggregator that fuses slow and fast temporal dynamics, and (3) a diagnostic output mechanism that enhances interpretability. Through multi-format training and hybrid supervision, Unified-VQA demonstrates state-of-the-art performance across 18 benchmarks, outperforming both conventional and deep learning-based metrics. Extensive ablation studies validate the contribution of each architectural and training component. Our results highlight the importance of format-aware learning and perceptual specialization in advancing generic VQA. Future work will explore interpretable multi-attribute diagnostics, lightweight deployment, and extensions to include user-generated content.


\bibliography{main}

\clearpage

\section{Supplementary}
\label{sec:Supplementary}

\begin{table*}[!t]
	\scriptsize
	\centering
	\caption{Performance of the proposed Unified-VQA (FR/NR) and benchmark approaches on eight Spatial-HD test databases.}
	\label{tab:results}
	\setlength{\tabcolsep}{5pt}
	\resizebox{\textwidth}{!}{\begin{tabular}{r|cccccccc|l}
		\toprule
		\textbf{SROCC}$\uparrow$ & \textbf{NFLX} & \textbf{NFLX-P} & \textbf{BVI-HD} & \textbf{CC-HD} & \textbf{CC-HDDO} & \textbf{MCL-V} & \textbf{SHVC} & \textbf{VQEGHD3} & {\textbf{Overall}} \\
		\midrule
		\multicolumn{10}{c}{\textbf{Full Reference (FR) Methods}}\\
		\midrule
		PSNR & 0.6218   & 0.6596   & 0.6143   & 0.6166   & 0.7497   & 0.4640   & 0.7380   & 0.7518   & 0.6520\\
		SSIM \cite{ssim} & 0.5638   & 0.6054   & 0.5992   & 0.7194   & 0.8026   & 0.4018   & 0.5446   & 0.7361   & 0.6216  \\
		MS-SSIM \cite{c:mssim} & 0.7136   & 0.7394   & 0.7652   & 0.7534   & 0.8321   & 0.6306   & 0.8007   & 0.8457   & 0.7601 \\
		DeepQA \cite{Kim_2017_CVPR} & 0.7298   & 0.6995   & 0.7106   & 0.6202   & 0.6705   & 0.6832   & 0.7176   & 0.7881   & 0.7024 \\
		LPIPS \cite{Lpips} & 0.6793   & 0.7859   & 0.6670   & 0.6838   & 0.7678   & 0.6579   & 0.6360   & 0.8075   & 0.7107 \\
		DeepVQA \cite{Kim2018DeepVQ} & 0.7352   & 0.7609   & 0.7330   & 0.6924   & 0.8120   & 0.6142   & 0.8041   & 0.7805   & 0.7540\\
		C3DVQA \cite{C3DVQA} & 0.7730   & 0.7714   & 0.7393   & 0.7203   & 0.8137   & 0.7126   & 0.8194   & 0.7329   & 0.7641 \\
		DISTS \cite{DISTS} & 0.7787   & 0.9325   & 0.7030   & 0.6303   & 0.7442   & 0.7792   & 0.7813   & 0.8254   & 0.7718 \\
		ST-GREED \cite{STGREED} & 0.7470   & 0.7445   & 0.7769   & 0.7738   & 0.8259   & 0.7226   & 0.7946   & 0.8079   & 0.7842 \\
		VMAF 0.6.1 \cite{w:VMAF} & 0.9254   & 0.9104   & 0.7962   & 0.8723   & 0.8783   & 0.7766   & 0.9114   & 0.8442   & 0.8644 \\
        RankDVQA (FR) \cite{feng2024rankdvqa} & 0.9393 & 0.9184 & 0.8991 & 0.9037 & 0.8659 & 0.8391 & 0.9142 & 0.8979 & 0.8972 \\\midrule 
        \textbf{Unified-VQA (FR)} & \textbf{0.9412} & \textbf{0.9238} & \textbf{0.9025} & \textbf{0.9051} & \textbf{0.9015} & \textbf{0.8459} & \textbf{0.9158} & \textbf{0.8991} & \textbf{0.9037} \\
		\midrule 
		\multicolumn{10}{c}{\textbf{No Reference (NR) Methods}}\\
		\midrule
		VIIDEO \cite{VIIDEO} & 0.4550   & 0.5527   & 0.1297   & 0.1308   & 0.2523   & 0.0406   & 0.2033   & 0.1881   & 0.2440 \\
		TLVQM \cite{TLVQM} & 0.4652   & 0.4720   & 0.3124   & 0.1622   & 0.3420   & 0.2758   & 0.4983   & 0.5382   & 0.3469\\
		BRISQUE \cite{brisque} & 0.7828   & 0.7861   & 0.2033   & 0.3738   & 0.3746   & 0.3154   & 0.3601   & 0.5467   & 0.4716 \\
		MDTVSFA \cite{MDTVSFA} & 0.5137   & 0.6024   & 0.3725   & 0.4068   & 0.5547   & 0.5712   & 0.6165   & 0.6422   & 0.5311 \\
		CONVIQT \cite{conviqt} & 0.6989   & 0.7962   & 0.3489   & 0.3706   & 0.5381   & 0.6323   & 0.4983   & 0.6217   & 0.5631 \\
		VIDEVAL \cite{tu2021ugc} & 0.7899   & 0.7261   & 0.5884   & 0.6974   & 0.7620   & 0.4836   & 0.6428   & 0.5326   & 0.6529  \\
		GSTVQA \cite{GSTVQA} & 0.8109   & 0.7858   & 0.4132   & 0.7447   & 0.7665   & 0.7385   & 0.6710   & 0.7011   & 0.7040 \\
        ChipQA \cite{Chipqa} & 0.7094   & 0.7395   & 0.6832   & 0.6533   & 0.7751   & 0.5425   & 0.6640   & 0.6954   & 0.6829 \\
		COVER \cite{he2024cover} & 0.8150   & 0.7911   & 0.7207   & 0.7273   & 0.7820   & 0.7192   & 0.7006   & 0.8009   & 0.7571  \\
        FastVQA \cite{FastVQA} & 0.8084   & 0.7738   & 0.7010   & 0.7585   & 0.7943   & 0.7450   & 0.7050   & 0.8210   & 0.7643  \\
        DOVER \cite{wu2023dover} & 0.8216 & 0.7855 & 0.7422 & 0.7629 & 0.8010 & 0.7324 & 0.7101 & 0.7932 & 0.7686 \\
        RankDVQA (NR) \cite{feng2024rankdvqa} & 0.8346 & 0.7944 & 0.7326 & 0.7628 & 0.7994 & 0.7631 & 0.7118 & 0.8346 & 0.7791 \\\midrule 
        \textbf{Unified-VQA (NR)} & \textbf{0.8355} & \textbf{0.8011} & \textbf{0.7913} & \textbf{0.7837} & \textbf{0.8182} & \textbf{0.7899} & \textbf{0.7304} & \textbf{0.8395} & \textbf{0.7987} \\
		\bottomrule
	\end{tabular}
    \label{tab:hd_results}}
\end{table*}

\section*{A. Evaluation Tasks and Scope}

In this experiment, to evaluate the effectiveness and generalization of the proposed Unified-VQA framework, we explicitly target modern professional streaming pipelines, focusing on various video formats (HD, UHD, HDR and HFR/VFI) associated with professionally generated content (PGC). While User-Generated Content (UGC) is another significant category, its inherent mixed distortions and lack of pristine references require a fundamentally different modeling approach \cite{qi2024full, qi2024bvi}. To ensure a rigorous evaluation of these format-specific artifacts, we restrict our scope to Professionally Generated Content (PGC) and leave the adaptation to the UGC domain for future research.

\section*{B. Additional Results: VQA Performance}
In the main paper, we reported the aggregated mean SROCC values for all four dataset groups (Spatial-HD, Spatial-UHD, Color, and Temporal). Here, we provide the detailed SROCC performance for each individual database in these four categories to further demonstrate the consistent performance of the proposed Unified-VQA. Furthermore, we have included five additional FR metrics: DeepQA \cite{Kim_2017_CVPR}, DeepVQA \cite{Kim2018DeepVQ}, C3DVQA \cite{C3DVQA}, DISTS \cite{DISTS}, and ST-GREED \cite{STGREED}; and eight NR metrics: VIIDEO \cite{VIIDEO}, TLVQM \cite{TLVQM}, BRISQUE \cite{brisque}, MDTVSFA \cite{MDTVSFA}, CONVIQT \cite{conviqt}, GSTVQA \cite{GSTVQA}, COVER \cite{he2024cover}, and DOVER \cite{wu2023dover}. We did not present their results because (i) the space in the main paper is limited and (ii) as shown in Table \ref{tab:results}, their overall correlation performance falls below that of the leading benchmarks (RankDVQA \cite{feng2024rankdvqa} and VMAF \cite{w:VMAF}).

\vspace{5pt}\noindent\textbf{Performance on Spatial-HD Databases.} Table \ref{tab:hd_results} details the performance on eight standard HD databases. It can be observed that Unified-VQA (FR) achieves the best performance across all eight datasets, consistently outperforming VMAF and RankDVQA, which rank second and third, in terms of overall performance. Moreover, the NR version of Unified-VQA also demonstrates superior performance, exhibiting a more balanced and robust performance across these databases, surpassing advanced NR quality metrics such as COVER \cite{he2024cover} and DOVER \cite{wu2023dover}.

\vspace{5pt}\noindent\textbf{Performance on Spatial-UHD, Color and Temporal datasets.}
Table \ref{tab:extended_results} presents the results on Spatial-UHD, Color and Temporal datasets. It is noted that the proposed Unified-VQA always offers the best performance among all tested quality models (in both FR and NR tracks) on all seven test sets. The only exceptional case is on the Yonsei-UHD database, where Unified-VQA-FR is second to VMAF-4K.

\section*{C. Additional Results: Artifact Detection}
While the main paper reports the F1-score, we provide additional results based on \textbf{Accuracy}  and \textbf{AUC}  on the Maxwell \cite{wu2023towards} (Table \ref{tab:maxvqa}) and BVI-Artifact \cite{feng2024bvi} (Table \ref{tab:bviartifact}) databases. These results further confirm the superior artifact detection performance of the proposed method.

\section*{D. Ablation Study based on Unified-VQA (NR)}
To further verify the effectiveness of each design component in our model, we conducted an additional ablation study on the Unified-VQA (NR) variant (the study results on the FR model are shown in the main paper). Similar to the full-reference analysis, we evaluated the contributions of the Perceptual Expert Network (PEN), the Spatio-Temporal Aggregator (STA), and the Unified Diagnostic Head (UDH). 

\vspace{5pt}\noindent \textbf{Effectiveness of the Perceptual Expert Network (PEN):} We replaced the domain-specific experts ($E_S, E_C, E_T$) with a single monolithic backbone to obtain \textbf{NR-V1}.

\vspace{5pt}\noindent \textbf{Effectiveness of the Spatio-Temporal Aggregator (STA):}
We replaced the SlowFast-based STA with a standard temporal pooling layer, resulting in \textbf{NR-V2}. 

\vspace{5pt}\noindent \textbf{Synergy of the Unified Diagnostic Head (UDH):} We implemented \textbf{NR-V3} by removing the diagnostic head and the auxiliary artifact loss.

These results in \autoref{tab:nr_ablation} show that all three variants are associated with inferior performance compared to the full NR Unified-VQA models - this verifies the contribution of the three key components in our design.

\begin{table*}[!t]
\centering
\caption{SROCC Performance of Unified-VQA (FR/NR) and benchmarks on Spatial-UHD, Color and Temporal databases.}
\label{tab:extended_results}
\resizebox{\textwidth}{!}{
\begin{tabular}{r|ccc|c|ccc}
\toprule
\textbf{SROCC} $\uparrow$ & \textbf{AVT-UHD} & \textbf{Yonsei-UHD} & \textbf{CC-UHD} & \textbf{LIVE HDR} & \textbf{BVI-VFI} & \textbf{BVI-HFR} & \textbf{LIVE-YT-HFR}\\
\midrule 
\multicolumn{8}{c}{\textbf{Full Reference (FR) Methods}} \\
\midrule
PSNR & 0.6825 & 0.8053 & 0.5517 & 0.3760 & 0.6532 & 0.3471 & 0.6950\\
SSIM \cite{ssim} & 0.7614 & 0.8184 & 0.5911 & 0.6976 & 0.6019 & 0.2802 & 0.4494\\
MS-SSIM \cite{c:mssim} & 0.7523 & 0.8890 & 0.7426 & 0.6757 & 0.6238 & 0.3187 & 0.4898\\
VMAF HD/4K \cite{w:VMAF} & 0.8891 & \textcolor{red}{0.9293} & 0.8463 & 0.8184 & 0.5841 & 0.1888 & 0.7782\\
LPIPS \cite{Lpips} & 0.7946 & 0.7396 & 0.8033 & 0.6826 & 0.5631 & 0.4182 & 0.6723 \\
HDR-VDP-3 \cite{hdrvdp3} & --- & --- & --- & 0.8080 & --- & --- & --- \\
RankDVQA (FR) \cite{feng2024rankdvqa} & 0.8901 & 0.8429 & 0.8319 & 0.8053 & 0.6039 & 0.6495 & 0.7492 \\ \midrule
\textbf{Unified-VQA (FR)} & \textbf{0.9037} & \textbf{0.9106} & \textbf{0.8560} & \textbf{0.8369} & \textbf{0.6653} & \textbf{0.7293} & \textbf{0.8123} \\
\midrule 
\multicolumn{8}{c}{\textbf{No Reference (NR) Methods}} \\
\midrule
VIDEVAL \cite{tu2021ugc} & 0.5529 & 0.6039 & 0.5616 & 0.4187 & 0.0830 & 0.3449 & 0.4742 \\
ChipQA \cite{Chipqa} & 0.6019 & 0.6788 & 0.5614 & 0.7435 & 0.3892 & 0.3092 & 0.4183\\
Fast-VQA \cite{FastVQA} & 0.6819 & 0.5456 & 0.4872 & 0.6538 & 0.1626 & 0.2875 & 0.4356 \\
RankDVQA (NR) \cite{feng2024rankdvqa} & 0.7302 & 0.7855 & 0.7691 & 0.7108 & 0.5122 & 0.4923 & 0.6659\\\midrule 
\textbf{Unified-VQA (NR)} & \textbf{0.7492} & \textbf{0.8094} & \textbf{0.7973} & \textbf{0.7524} & \textbf{0.5658} & \textbf{0.6874} & \textbf{0.7038} \\
\bottomrule
\end{tabular}
}
\end{table*}

\begin{table*}[!t]
    \centering
    \caption[Artifact detection results on the Maxwell database.]{Artifact detection results on the Maxwell database~\cite{wu2023towards}. Here `\colorbox{mygray}{ - }' indicates that the tested method in this row is not designed to identify the corresponding artifact in this column.\label{tab:maxvqa}}
    \resizebox{\linewidth}{!}{\begin{tabular}{c r c c c c c c c c c c}
    \toprule

       Metric  & Method  & Motion & Dark (Lighting) & Grain.(Noise) & Block. (Compression) & Spat.(Focus) & Drop.\\
       \midrule
      \multirow{7}{*}{Acc. (\%) $\uparrow$} 
       & EFENet \cite{zhao2021defocus} &\cellcolor[HTML]{EFEFEF}- &\cellcolor[HTML]{EFEFEF}- &\cellcolor[HTML]{EFEFEF}- &\cellcolor[HTML]{EFEFEF}- & 54.55 &\cellcolor[HTML]{EFEFEF}- \\
      & MLDBD \cite{zhao2023full} &\cellcolor[HTML]{EFEFEF}- &\cellcolor[HTML]{EFEFEF}- &\cellcolor[HTML]{EFEFEF}- &\cellcolor[HTML]{EFEFEF}- & 56.38 &\cellcolor[HTML]{EFEFEF}- \\
      & Wolf \textit{et al.} \cite{wolf2008no} &\cellcolor[HTML]{EFEFEF}- &\cellcolor[HTML]{EFEFEF}- &\cellcolor[HTML]{EFEFEF}- &\cellcolor[HTML]{EFEFEF}- &\cellcolor[HTML]{EFEFEF}- & 50.63 \\
      & VIDMAP \cite{vidmap} &\cellcolor[HTML]{EFEFEF}- &\cellcolor[HTML]{EFEFEF}- &\cellcolor[HTML]{EFEFEF}- & 70.90 & 58.96 & 59.20 \\
      & MaxVQA~\cite{wu2023towards} & 78.30 & 75.69 & 65.10 & 85.42 & 82.66 & 79.66 \\
      & MVAD \cite{feng2025mvad} & 82.64 & 79.91 & 72.67 & \textbf{98.00} & \textbf{90.46} & \textbf{81.68} \\ 
      \cmidrule{2-8}
      & \textbf{Unified-VQA} & \textbf{85.24} & \textbf{80.65} & \textbf{74.89} & \textbf{98.00} & \textbf{90.46} & \textbf{81.68} \\
      \midrule
      \multirow{7}{*}{F1 $\uparrow$} 
      & EFENet \cite{zhao2021defocus} &\cellcolor[HTML]{EFEFEF}- &\cellcolor[HTML]{EFEFEF}- &\cellcolor[HTML]{EFEFEF}- &\cellcolor[HTML]{EFEFEF}- & 0.66 &\cellcolor[HTML]{EFEFEF}- \\
      & MLDBD \cite{zhao2023full} &\cellcolor[HTML]{EFEFEF}- &\cellcolor[HTML]{EFEFEF}- &\cellcolor[HTML]{EFEFEF}- &\cellcolor[HTML]{EFEFEF}- & 0.63 &\cellcolor[HTML]{EFEFEF}- \\
      & Wolf \textit{et al.} \cite{wolf2008no} &\cellcolor[HTML]{EFEFEF}- &\cellcolor[HTML]{EFEFEF}- &\cellcolor[HTML]{EFEFEF}- &\cellcolor[HTML]{EFEFEF}- &\cellcolor[HTML]{EFEFEF}- & 0.59 \\
      & VIDMAP \cite{vidmap} &\cellcolor[HTML]{EFEFEF}- &\cellcolor[HTML]{EFEFEF}- &\cellcolor[HTML]{EFEFEF}- & 0.71 & 0.64 & 0.62 \\
      & MaxVQA~\cite{wu2023towards} & 0.75 & 0.72 & 0.66 & 0.85 & 0.82 & 0.77 \\
      &  MVAD \cite{feng2025mvad} & 0.78 & 0.80 & 0.72 & \textbf{0.99} & \textbf{0.92} & \textbf{0.79} \\ 
       \cmidrule{2-8}
      & \textbf{Unified-VQA}  & \textbf{0.80}         & \textbf{0.82}         & \textbf{0.75}        & \textbf{0.99}         & \textbf{0.92}         & \textbf{0.79} \\
      \midrule
      \multirow{7}{*}{AUC $\uparrow$} 
      & EFENet \cite{zhao2021defocus} &\cellcolor[HTML]{EFEFEF}- &\cellcolor[HTML]{EFEFEF}- &\cellcolor[HTML]{EFEFEF}- &\cellcolor[HTML]{EFEFEF}- & 0.63 &\cellcolor[HTML]{EFEFEF}- \\
      & MLDBD \cite{zhao2023full} &\cellcolor[HTML]{EFEFEF}- &\cellcolor[HTML]{EFEFEF}- &\cellcolor[HTML]{EFEFEF}- &\cellcolor[HTML]{EFEFEF}- & 0.61 &\cellcolor[HTML]{EFEFEF}- \\
      & Wolf \textit{et al.} \cite{wolf2008no} &\cellcolor[HTML]{EFEFEF}- &\cellcolor[HTML]{EFEFEF}- &\cellcolor[HTML]{EFEFEF}- &\cellcolor[HTML]{EFEFEF}- &\cellcolor[HTML]{EFEFEF}- & 0.62 \\
      & VIDMAP \cite{vidmap} &\cellcolor[HTML]{EFEFEF}- &\cellcolor[HTML]{EFEFEF}- &\cellcolor[HTML]{EFEFEF}- & 0.70 & 0.62 & 0.64 \\
      & MaxVQA~\cite{wu2023towards} & 0.81 & 0.76 & 0.63 & 0.88 & 0.87 & 0.78 \\
      & MVAD \cite{feng2025mvad} & 0.85 & 0.78 & 0.68 & \textbf{0.99} & 0.90 & 0.78 \\     
       \cmidrule{2-8}
      & \textbf{Unified-VQA} & \textbf{0.86}         & \textbf{0.81}         & \textbf{0.71}        & \textbf{0.99}         & \textbf{0.92}         & \textbf{0.79} \\
      \bottomrule
    \end{tabular}}
\end{table*}

\begin{table*}[!t]
    \centering
    \caption[Artifact detection results on the BVI-Artifact database.]{Artifact detection results on the BVI-Artifact database \cite{feng2024bvi}. Here `\colorbox{mygray}{ - }' indicates that the tested method in this row is not designed to identify the corresponding artifact in this column. \label{tab:bviartifact}}
    \resizebox{1\linewidth}{!}{\begin{tabular}{c r c c c c c c c c c c}
    \toprule
       Metric  & Method  & Motion & Dark & Grain. & Alias. & Band. & Block. & Spat. & Drop. & Trans. & Black.\\
       \midrule
      \multirow{9}{*}{Acc. (\%) $\uparrow$} 
      & CAMBI \cite{tandon2021cambi} & \cellcolor[HTML]{EFEFEF}- & \cellcolor[HTML]{EFEFEF}- & \cellcolor[HTML]{EFEFEF}- & \cellcolor[HTML]{EFEFEF}- & 61.88 & \cellcolor[HTML]{EFEFEF}- & \cellcolor[HTML]{EFEFEF}- & \cellcolor[HTML]{EFEFEF}- & \cellcolor[HTML]{EFEFEF}- & \cellcolor[HTML]{EFEFEF}- \\
      & BBAND \cite{tu2020bband} & \cellcolor[HTML]{EFEFEF}- & \cellcolor[HTML]{EFEFEF}- & \cellcolor[HTML]{EFEFEF}- & \cellcolor[HTML]{EFEFEF}- & 50.00 & \cellcolor[HTML]{EFEFEF}- & \cellcolor[HTML]{EFEFEF}- & \cellcolor[HTML]{EFEFEF}- & \cellcolor[HTML]{EFEFEF}- & \cellcolor[HTML]{EFEFEF}- \\
      & EFENet \cite{zhao2021defocus} & \cellcolor[HTML]{EFEFEF}- & \cellcolor[HTML]{EFEFEF}- & \cellcolor[HTML]{EFEFEF}- & \cellcolor[HTML]{EFEFEF}- & \cellcolor[HTML]{EFEFEF}- & \cellcolor[HTML]{EFEFEF}- & 47.08 & \cellcolor[HTML]{EFEFEF}- & \cellcolor[HTML]{EFEFEF}- & \cellcolor[HTML]{EFEFEF}- \\
      & MLDBD \cite{zhao2023full} & \cellcolor[HTML]{EFEFEF}- & \cellcolor[HTML]{EFEFEF}- & \cellcolor[HTML]{EFEFEF}- & \cellcolor[HTML]{EFEFEF}- & \cellcolor[HTML]{EFEFEF}- & \cellcolor[HTML]{EFEFEF}- & 49.58 & \cellcolor[HTML]{EFEFEF}- & \cellcolor[HTML]{EFEFEF}- & \cellcolor[HTML]{EFEFEF}- \\
      & Wolf \textit{et al.} \cite{wolf2008no} & \cellcolor[HTML]{EFEFEF}- & \cellcolor[HTML]{EFEFEF}- & \cellcolor[HTML]{EFEFEF}- & \cellcolor[HTML]{EFEFEF}- & \cellcolor[HTML]{EFEFEF}- & \cellcolor[HTML]{EFEFEF}- & \cellcolor[HTML]{EFEFEF}- & 51.67 & \cellcolor[HTML]{EFEFEF}- & \cellcolor[HTML]{EFEFEF}- \\
      & VIDMAP \cite{vidmap} & \cellcolor[HTML]{EFEFEF}- & \cellcolor[HTML]{EFEFEF}- & \cellcolor[HTML]{EFEFEF}- & 50.00 & 56.25 & 54.38 & 47.29 & 45.42 & 51.04 & \cellcolor[HTML]{EFEFEF}- \\
      & MaxVQA~\cite{wu2023towards} & 51.88 & 73.13 & 38.75 & \cellcolor[HTML]{EFEFEF}- & \cellcolor[HTML]{EFEFEF}- & 64.58 & 53.54 & \cellcolor[HTML]{EFEFEF}- & \cellcolor[HTML]{EFEFEF}- & \cellcolor[HTML]{EFEFEF}- \\
      & MVAD \cite{feng2025mvad} & 59.38 & 78.32 & 66.87 & 93.75 & 64.96 & \textbf{99.97} & \textbf{92.92} & 71.25 & 74.17 & 76.25 \\ 
       \cmidrule{2-12}
      & \textbf{Unified-VQA} & \textbf{62.04} & \textbf{80.45} & \textbf{69.29} & \textbf{95.25} & \textbf{67.53} & \textbf{99.97} & \textbf{92.92} & \textbf{73.76} & \textbf{75.85} & \textbf{78.25} \\
      \midrule
      \multirow{9}{*}{F1 $\uparrow$} 
      & CAMBI \cite{tandon2021cambi} & \cellcolor[HTML]{EFEFEF}- & \cellcolor[HTML]{EFEFEF}- & \cellcolor[HTML]{EFEFEF}- & \cellcolor[HTML]{EFEFEF}- & 0.53 & \cellcolor[HTML]{EFEFEF}- & \cellcolor[HTML]{EFEFEF}- & \cellcolor[HTML]{EFEFEF}- & \cellcolor[HTML]{EFEFEF}- & \cellcolor[HTML]{EFEFEF}- \\
      & BBAND \cite{tu2020bband} & \cellcolor[HTML]{EFEFEF}- & \cellcolor[HTML]{EFEFEF}- & \cellcolor[HTML]{EFEFEF}- & \cellcolor[HTML]{EFEFEF}- & 0.44 & \cellcolor[HTML]{EFEFEF}- & \cellcolor[HTML]{EFEFEF}- & \cellcolor[HTML]{EFEFEF}- & \cellcolor[HTML]{EFEFEF}- & \cellcolor[HTML]{EFEFEF}- \\
      & EFENet \cite{zhao2021defocus} & \cellcolor[HTML]{EFEFEF}- & \cellcolor[HTML]{EFEFEF}- & \cellcolor[HTML]{EFEFEF}- & \cellcolor[HTML]{EFEFEF}- & \cellcolor[HTML]{EFEFEF}- & \cellcolor[HTML]{EFEFEF}- & 0.64 & \cellcolor[HTML]{EFEFEF}- & \cellcolor[HTML]{EFEFEF}- & \cellcolor[HTML]{EFEFEF}- \\
      & MLDBD \cite{zhao2023full} & \cellcolor[HTML]{EFEFEF}- & \cellcolor[HTML]{EFEFEF}- & \cellcolor[HTML]{EFEFEF}- & \cellcolor[HTML]{EFEFEF}- & \cellcolor[HTML]{EFEFEF}- & \cellcolor[HTML]{EFEFEF}- & 0.65 & \cellcolor[HTML]{EFEFEF}- & \cellcolor[HTML]{EFEFEF}- & \cellcolor[HTML]{EFEFEF}- \\
      & Wolf \textit{et al.} \cite{wolf2008no} & \cellcolor[HTML]{EFEFEF}- & \cellcolor[HTML]{EFEFEF}- & \cellcolor[HTML]{EFEFEF}- & \cellcolor[HTML]{EFEFEF}- & \cellcolor[HTML]{EFEFEF}- & \cellcolor[HTML]{EFEFEF}- & \cellcolor[HTML]{EFEFEF}- & 0.18 & \cellcolor[HTML]{EFEFEF}- & \cellcolor[HTML]{EFEFEF}- \\
      & VIDMAP \cite{vidmap} & \cellcolor[HTML]{EFEFEF}- & \cellcolor[HTML]{EFEFEF}- & \cellcolor[HTML]{EFEFEF}- & 0.67 & 0.59 & 0.69 & 0.64 & 0.59 & 0.65 & \cellcolor[HTML]{EFEFEF}- \\
      & MaxVQA~\cite{wu2023towards} & 0.68 & 0.67 & 0.16 & \cellcolor[HTML]{EFEFEF}- & \cellcolor[HTML]{EFEFEF}- & 0.55 & 0.40 & \cellcolor[HTML]{EFEFEF}- & \cellcolor[HTML]{EFEFEF}- & \cellcolor[HTML]{EFEFEF}- \\
      & MVAD \cite{feng2025mvad} & 0.69 & 0.73 & 0.46 & 0.90 & 0.64 & \textbf{0.99} & 0.90 & 0.65 & 0.73 & 0.65 \\ 
       \cmidrule{2-12}
      & \textbf{Unified-VQA} & \textbf{0.74} & \textbf{0.80} & \textbf{0.58} & \textbf{0.93} & \textbf{0.68} & \textbf{0.99} & \textbf{0.92} & \textbf{0.71} & \textbf{0.75} & \textbf{0.68} \\
      \midrule
      \multirow{9}{*}{AUC $\uparrow$} 
      & CAMBI \cite{tandon2021cambi} & \cellcolor[HTML]{EFEFEF}- & \cellcolor[HTML]{EFEFEF}- & \cellcolor[HTML]{EFEFEF}- & \cellcolor[HTML]{EFEFEF}- & 0.63 & \cellcolor[HTML]{EFEFEF}- & \cellcolor[HTML]{EFEFEF}- & \cellcolor[HTML]{EFEFEF}- & \cellcolor[HTML]{EFEFEF}- & \cellcolor[HTML]{EFEFEF}- \\
      & BBAND \cite{tu2020bband} & \cellcolor[HTML]{EFEFEF}- & \cellcolor[HTML]{EFEFEF}- & \cellcolor[HTML]{EFEFEF}- & \cellcolor[HTML]{EFEFEF}- & 0.51 & \cellcolor[HTML]{EFEFEF}- & \cellcolor[HTML]{EFEFEF}- & \cellcolor[HTML]{EFEFEF}- & \cellcolor[HTML]{EFEFEF}- & \cellcolor[HTML]{EFEFEF}- \\
      & EFENet \cite{zhao2021defocus} & \cellcolor[HTML]{EFEFEF}- & \cellcolor[HTML]{EFEFEF}- & \cellcolor[HTML]{EFEFEF}- & \cellcolor[HTML]{EFEFEF}- & \cellcolor[HTML]{EFEFEF}- & \cellcolor[HTML]{EFEFEF}- & 0.57 & \cellcolor[HTML]{EFEFEF}- & \cellcolor[HTML]{EFEFEF}- & \cellcolor[HTML]{EFEFEF}- \\
      & MLDBD \cite{zhao2023full} & \cellcolor[HTML]{EFEFEF}- & \cellcolor[HTML]{EFEFEF}- & \cellcolor[HTML]{EFEFEF}- & \cellcolor[HTML]{EFEFEF}- & \cellcolor[HTML]{EFEFEF}- & \cellcolor[HTML]{EFEFEF}- & 0.53 & \cellcolor[HTML]{EFEFEF}- & \cellcolor[HTML]{EFEFEF}- & \cellcolor[HTML]{EFEFEF}- \\
      & Wolf \textit{et al.} \cite{wolf2008no} & \cellcolor[HTML]{EFEFEF}- & \cellcolor[HTML]{EFEFEF}- & \cellcolor[HTML]{EFEFEF}- & \cellcolor[HTML]{EFEFEF}- & \cellcolor[HTML]{EFEFEF}- & \cellcolor[HTML]{EFEFEF}- & \cellcolor[HTML]{EFEFEF}- & 0.60 & \cellcolor[HTML]{EFEFEF}- & \cellcolor[HTML]{EFEFEF}- \\
      & VIDMAP \cite{vidmap} & \cellcolor[HTML]{EFEFEF}- & \cellcolor[HTML]{EFEFEF}- & \cellcolor[HTML]{EFEFEF}- & 0.58 & 0.58 & 0.61 & 0.38 & 0.47 & 0.50 & \cellcolor[HTML]{EFEFEF}- \\
      & MaxVQA~\cite{wu2023towards} & 0.56 & \textbf{0.84} & 0.36 & \cellcolor[HTML]{EFEFEF}- & \cellcolor[HTML]{EFEFEF}- & 0.80 & 0.54 & \cellcolor[HTML]{EFEFEF}- & \cellcolor[HTML]{EFEFEF}- & \cellcolor[HTML]{EFEFEF}- \\
      & MVAD \cite{feng2025mvad} & 0.60 & 0.78 & 0.56 & 0.93 & 0.67 & \textbf{0.99} & \textbf{0.93} & 0.71 & 0.80 & 0.50 \\ 
       \cmidrule{2-12}
      & \textbf{Unified-VQA} & \textbf{0.62} & 0.80 & \textbf{0.57} & \textbf{0.95} & \textbf{0.68} & \textbf{0.99} & \textbf{0.93} & \textbf{0.73} & \textbf{0.81} & \textbf{0.54} \\
      \bottomrule
    \end{tabular}}

\end{table*}

\begin{table*}[!t]
\centering
\caption{Ablation study results for the proposed Unified-VQA (NR) variant. We evaluate the contribution of the Perceptual Expert Network (PEN), the Spatio-Temporal Aggregator (STA), and the Unified Diagnostic Head (UDH) in the absence of reference information.}
\label{tab:nr_ablation}
\resizebox{0.85\linewidth}{!}{\begin{tabular}{l|c|c|c|c}
\toprule
\textbf{Average SROCC} $\uparrow$ & \textbf{Spatial-HD (8)} & \textbf{Spatial-UHD (3)} & \textbf{Color (1)} & \textbf{Temporal (3)} \\
\midrule 
\textbf{NR-V1} (Single expert Net) & 0.7512$\pm$0.0510 & 0.7348$\pm$0.0395 & 0.7005 & 0.5984$\pm$0.0892 \\
\textbf{NR-V2} (CNN Aggregator) & 0.7805$\pm$0.0462 & 0.7652$\pm$0.0342 & 0.7288 & 0.5612$\pm$0.1154 \\
\textbf{NR-V3} (Without UDH) & 0.7892$\pm$0.0435 & 0.7724$\pm$0.0320 & 0.7356 & 0.6245$\pm$0.0841 \\
\midrule
\textbf{Unified-VQA (NR)} & \textbf{0.7987$\pm$0.0426} & \textbf{0.7853$\pm$0.0311} & \textbf{0.7524} & \textbf{0.6523$\pm$0.0724} \\
\bottomrule
\end{tabular}}
\end{table*}

\section*{E. Limitations}
\label{sec:limitations}
While Unified-VQA has demonstrated great model generalization and consistent performance improvement over existing methods, it is associated with the following limitations. First, the Perceptual Expert Network (PEN) focuses specifically on spatial, color, and temporal domains; while effective for standard pipelines, it does not yet address other specialized artifacts with formats such as VR or audio (audio-video synchronization) - our flexible PEN architecture does support such extensions though. Secondly, the model primarily assesses low-level video fidelity, lacking the high-level semantic understanding required to differentiate between artistic aesthetics (e.g., intentional bokeh) and technical errors. Finally, our current scope is restricted to Professionally Generated Content (PGC); future work should extend the framework to address the complex, commingled distortions found in User-Generated Content (UGC) and application-specific restoration artifacts.

\section*{F. Social Impact}
\label{sec:social_impact}

\textbf{Environmental Sustainability.} The deployment of deep learning-based VQA models inevitably incurs a higher computational cost compared to traditional pixel-wise distortion-based metrics (e.g., PSNR, SSIM), leading to increased energy consumption and carbon emissions during training and inference. However, compared to VQA methods based on Large Multimodal Models (LMMs), Unified-VQA is much more lightweight and energy-efficient. By achieving excellent VQA performance with just a fraction of the parameters required by LMMs, our approach represents an excellent trade off between performance and environmental cost.

\textbf{Improving Digital Experience.} On the positive side, Unified-VQA could potentially serve as a critical tool for the video streaming industry. By providing accurate, diagnostic quality feedback, it enables service providers to monitor video quality and optimize encoding parameters. This ensures that billions of end-users receive a superior Quality of Experience (QoE) while potentially saving bandwidth by avoiding over-allocating bitrates to content that is already perceptually high-quality.
\end{document}